\documentclass{article}

\usepackage{arxiv}

\usepackage[utf8]{inputenc} 
\usepackage[T1]{fontenc}    
\usepackage{hyperref}       
\usepackage{url}            
\usepackage{booktabs}       
\usepackage{amsfonts}       
\usepackage{nicefrac}       
\usepackage{microtype}      
\usepackage{lipsum}
\usepackage{verbatim} 
\usepackage{amsmath}
\usepackage{graphicx}

\title{\emph{Two-phase flow regime prediction using LSTM based deep recurrent neural network}}

\author{
 Zhuoran Dang\thanks{email: zdang@purdue.edu} \\
 School of Nuclear Engineering\\
  Purdue University\\
  West Lafayette, IN 47907 \\
    \And
 Mamoru Ishii \\
  School of Nuclear Engineering\\
  Purdue University\\
  West Lafayette, IN 47907 \\}

\begin{document}
\maketitle

\begin{abstract}
Long short-term memory (LSTM) and recurrent neural network (RNN) has achieved great successes on time-series prediction. In this paper, a methodology of using LSTM-based deep-RNN for two-phase flow regime prediction is proposed, motivated by previous research on constructing deep RNN. The method is featured with fast response and accuracy. The built RNN networks are trained and tested with time-series void fraction data collected using impedance void meter. The result shows that the prediction accuracy depends on the depth of network and the number of layer cells. However, deeper and larger network consumes more time in predicting. 
\end{abstract}


\section{Introduction}
Two-phase flow regime is an important concept for severe accident prediction and prevention in two-phase flow systems such as the reactor pressure vessel in the nuclear power plant. It serves as an engineering reference that classifies flow characteristics. Many two-phase flow models are based on flow regimes. Thus, an accurate prediction on flow regime can be regarded as the first step towards an accurate two-phase flow prediction. The analysis of the two-phase flow regime and its transitions has quite a long history. Flow regime maps were developed for different flow geometries \cite{mishima1984flow, mishima1996some, barnea1983flow}. Flow regimes are determined by using two-phase parameters that can be experimentally obtained. In the early times, the flow regime maps are either created using experimental data or theoretical approaches \cite{mishima1984flow}. Recently, the flow regime identifications are developed with the help of machine learning techniques.

Over the past years, exclusive work has been done on two-phase flow regime identification using machine learning algorithms. Among them, supervised multi-layer feedforward neural networks (NN), supported vector machine (SVM) \cite{cortes1995support} and self-organized map (SOM) \cite{kohonen1990self} are the most used algorithms and other generated algorithms are more or less based on these algorithms. \cite{mi1998vertical, dang2019investigation, yunlong2008identification} They basically require to input all the data to extract the different key features and make predictions. Although these methods are proved to be accurate, certain shortcomings of these methods are: 1) most of the algorithms are determining static flow regime maps that are used as engineering references, while the structures and two-phase flow is changing dynamically; 2) the flow regime are only be determined afterwards. However, severe accidents, such as nuclear power plant core melting, often happens at sudden, a fast response and accident prediction is essentially needed. 
In this paper, a dynamic RNN approach is proposed for the two-phase flow regime prediction. A LSTM-based deep RNN is constructed and trained using existing database and the performance is evaluated and analyzed in this paper. 

\section{Recurrent Neural Network}
\label{sec:headings}

RNN is structurally suitable for a time-series prediction. Conventional RNN can process time series data temporally and dynamically based on hidden Markov model (HMM), which makes RNN to be able to capture long-distance dependencies. However, RNN could fall into the trouble of gradient vanishing and exploring during model training. Long short-term memory (LSTM) networks \cite{hochreiter1997long} was created and solved this problem properly by managing the passes of the information. In a LSTM system, the recurrent hidden layers are computing with the self-connected memory cells and three gates for obtaining the outputs. The key of LSTM in solving the gradient vanishing and exploring is to optionally ignore some of the inputs so that they aren't used for the updates of parameters in the hidden states.

Given a time-dependent void fraction sequence $\alpha$ = {[$\alpha_1$, $\alpha_2$, $\alpha_3$, ..., $\alpha_n$]}, the mathematical expressions for the operation of one LSTM hidden cell t, are given as follows,
\begin{equation}
\begin{aligned}
  i_{t} =\sigma\left(W_{\alpha i} \alpha_{t}+W_{h i} h_{t-1}+W_{c i} c_{t-1}+b_{i}\right) \\  
  f_{t} =\sigma\left(W_{\alpha f} x_{t}+W_{h f} h_{t-1}+W_{c f} c_{t-1}+b_{f}\right) \\   
  a_{t} =\tau\left(W_{\alpha c} \alpha_{t}+W_{h c} h_{t-1}+b_{c}\right) \\  
  c_{t} =f_{t} c_{t-1}+i_{t} a_{t} \\  
  o_{t} =\sigma\left(W_{\alpha o} \alpha_{t}+W_{h o} h_{t-1}+W_{c o} c_{t}+b_{o}\right) \\  
  h_{t} =o_{t} \theta\left(c_{t}\right)
\end{aligned}
\end{equation}

where \emph{W} is the weights for each parameter at certain state. \emph{ht} is the vector sequence of the hidden cell. In this model, the activation functions are sigmoid function, $\sigma$, and tanh function, $\tau$. A LSTM system usually contains multiple connected cells among which the outputs from the preceding cell are the inputs of the following cell. The characteristics of the two phase flow is able to pass thought the model. The output of the model is a probability distribution of all the possible flow regimes. The predicted flow regime is the one with the highest probability, obtained as follows,

The prediction on flow regime using LSTM-based RNN has its advantages. Firstly, the input sequence is segmented. Each input node in the sequence represents the state of the flow regime at certain time. Secondly, the relation between the sequence and the output is rather tight since the sequence hardly contains unrelated noises. However, since the mechanisms in related with the transition and development of the two phase flow regime is complicated and the number of dependencies is large, a deep LSTM network is still needed.
  
This paper follows a similar approach of constructing deep LSTM network with \cite{li2015constructing}. Their ideas of constructing deep RNN network is as follows: 1) input-hidden; 2) hidden-hidden; 3) hidden-output. Based on the ideas, 5 different types of RNN network are constructed by combining different sublayers. In terms of constructing the network, one important consideration is that with the basis of accuracy, the network should respond quickly. This means that the setting of the number of the parameters should balance both the requirement of accuracy and the calculation latency.

\begin{figure}[h]
\begin{center}
  \includegraphics[width=0.6\linewidth]{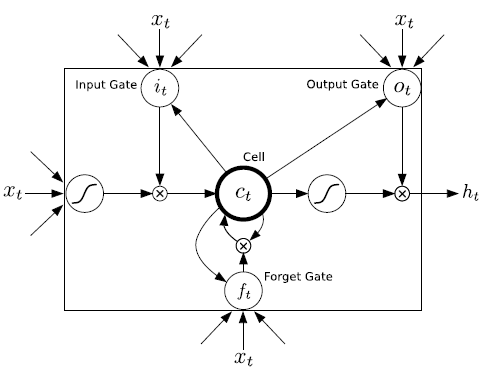}
  \caption{Structure of LSTM cell \cite{graves2013speech}}
  \label{fig:fig1}
\end{center}
\end{figure}

\begin{figure}[h]
\begin{center}
  \includegraphics[width=0.6\linewidth]{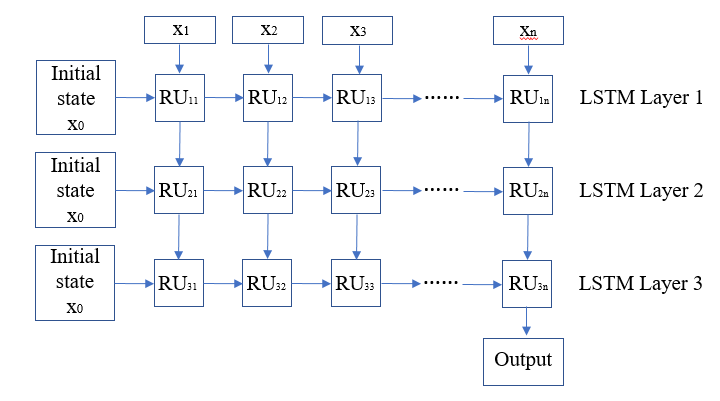}
  \caption{Example of structure of 3 LSTM-RNN hidden layers}
  \label{fig:fig2}
\end{center}
\end{figure}

\section{Experiments}
\label{sec:headings}

We evaluated the LSTM-based D-RNN on the current existing time-series void fraction signal database  \cite{dang2019investigation}. The details on experiment setup, database, and the experimental results are discussed below.

\subsection{Database}
The two-phase flow regime parameters that can describe the flow regime could be classified into two groups. The parameters in the first group directly describe the flow regime characteristics, such as void fraction and interfacial area concentration. The parameters in the second group also include the flow regime characteristics such as local pressure in the system. Since parameters in both groups include the characteristics of flow regime, they could be used as the input parameters for the flow regime prediction.

Time series data containing two-phase flow regime characteristics can be obtained using many two-phase flow measurement instrumentations. In the lab setup, gamma densitometer is a very accurate and stable method because it is non-intrusive, almost flow-regime dependence-free instrumentation. \cite{pune9203} Conductivity probe is another accurate instrumentation yet its setup is relatively difficult. Impedance void meter, as an engineering reference, is also non-intrusive yet its accuracy dependence on the flow regime and void distribution.\cite{hewitt1978measurement} In terms of industrial application, differential pressure gauge is a convenient and economical choice, yet its accuracy of measurement is not satisfying. All of these mentioned above can provide the time series data that contains the void fraction changing characteristics. The database used in this experiment \cite{dang2019investigation} is collected using impedance meter. 

\begin{figure}[h]
\begin{center}
  \includegraphics[width=0.8\linewidth]{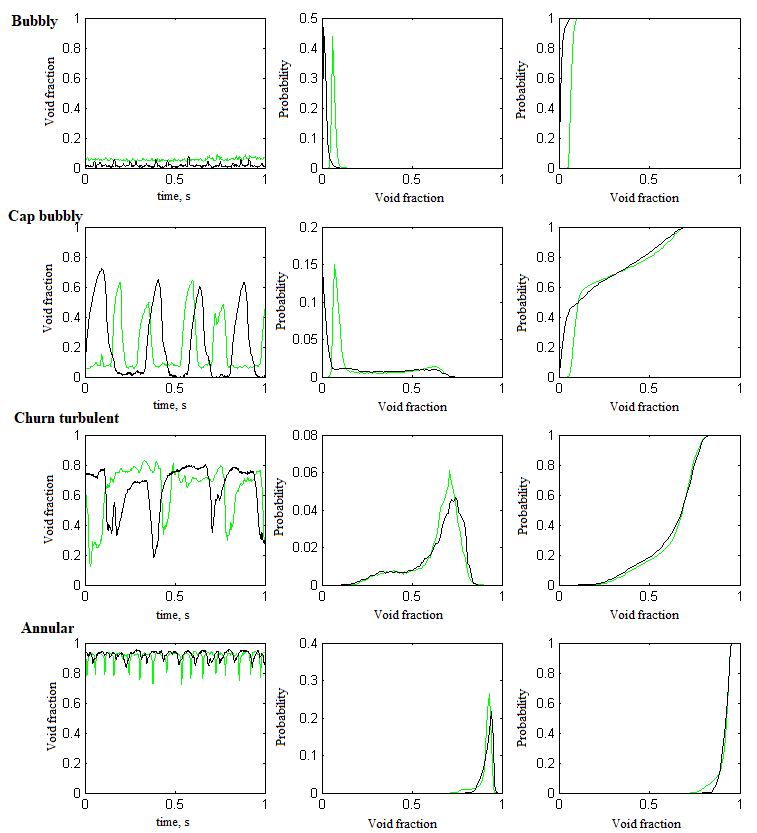}
  \caption{Example of time series void fraction signal (left column), probability density function (middle column), and cumulative probability density function (right column) for different flow regimes. In each figure, the black and green curves represent two different test cases belonging to one flow regime. Data from \cite{dang2019investigation}.}
  \label{fig:fig3}
\end{center}
\end{figure}

In the database, flow regime is classified into 5 types including bubbly, cap bubbly, slug, churn-turbulent, and annular flow. The database contains 200 test conditions in total, and each test condition consists of an impedance signal with a measurement period of 60 seconds and data acquisition frequency of 10kHz. For each test condition, signal is ranging from 0 to 1, with 0 representing full water case and 1 representing full air case, signal fluctuating between 0 and 1 according to the flow regime characteristics. \ref{fig:fig3} shows the 1 second time signal, probability density functions (PDF) and the cumulative probability density functions (CPDF) to characterize each flow regime. The PDF or CPDF profiles are usually treated as inputs in SOM or SVM methods.

Data augmentation was performed with the original database. Two methods are used in this experiment and they are summarized below.

\begin{itemize}
\item	Since the experiment was performed at steady-state conditions, meaning that the flow regime state is not changing during data collection, data could be segmented into shorter pieces. In the following data sensitivity analysis part, the performance in terms of the length of the data is evaluated and discussed.
\item	The time-series signals are reversed in order. In this way, the number of data is doubled.
\end{itemize}

\subsection{Experimental setup}
The void fraction signal is firstly put into one ReLU-NN layer for the primary feature extraction. The ReLU-NN layer (or ReLU layer) stands for feedforward neural network layer with ReLU as the activation function. Then it passes through LSTM layers and ReLU layers. The structures of these parts are modified and investigated for the better prediction. The final output stage of the network is a softmax layer with a size of the number of all possible flow regime. The networks are established with tensorflow-based Keras. The optimizer used for all the networks in the experiment is Adam and the loss function is categorical cross-entropy.

For the model training, although LSTM unit is featured with good preventing of gradient vanishing and exploration, creating and training and good LSTM-based RNN model require good methods and tricks. Some of the important methodologies used towards building and training a well-established RNN model are utilized during the model training.
\begin{itemize}
\item Gate weight initiation: Gates are the keys for a LSTM unit. Gate initial states can have large effect on the training process. It has been proved that deep neural networks can converge more rapidly using orthogonal initiation that generates a random orthogonal matrix.\cite{saxe2013exact} 
\item Dropout and regularization: To prevent from over-fitting and better training the model, the following regularization methods are used: early stopping and learning rate reduction. The training time is largely decreased due to these two methods. By using early stopping, the training process will stop if testing accuracy has not been improved for 3 epochs. With learning rate reduction, the learning rate is initialized as 0.01 and it can decease according to the training performance. The minimum learning rate allowed is 0.0001. 
\end{itemize}

\subsection{Sensitivity Study}

The length of the time-series data can affect the flow regime prediction. This is obvious because a very short sequence may not include all the key information of the characteristics of the two phase flow. However, if the sequence is very long, the prediction will not be a "real-time" prediction. Therefore, it is essential to determine the proper length of the sequence for the flow regime prediction.    

Different flow regime contains different characteristics and the sequence length needed for each characteristic to be presented is different. From \ref{fig:fig3}, the signal variations of typical bubbly and annular two-phase flow is small and the PDF curve contains only one peak. This means that the characteristics of these two-phase flows are uniform over the time and a short sequence could present these characteristics. In contrast, the signal fluctuation of the flow regimes like slug and churn-turbulent two-phase flow usually quite large. Thus, these flow regimes usually require a longer sequence. Since our objective is to classify the flow regimes, the length of sequence is determined by the flow regime that requires the longest sequence. Besides, other hydrodynamic parameters such as flow rate should also be considered when determining the sequence length. It is obvious that flow rate determines the speed of the two-phase flow passing through the measurement area, thus affecting the sequence length needed.

In this study, the experimental data are segmented into different sequence lengths and they are used separately for training the same model structure. The lengths and the training performances are given in table ~\ref{tab:seqlen}. The selection of these lengths is by considering the types of flow regimes and test conditions included in the database. The model structure used in this section, LTSM-2ReLU,
From the table, the prediction accuracy of the model increases as the sequence length becomes longer. In terms of the performance tendency, the performance of the model trained using data with 3 seconds sequence length drastically worsen compared with result of 5 seconds. This may not be a general conclusion but it provides a method of determining the input data size. 

\begin{table}[h]
 \caption{Sensitivity study on the effect of sequence length}
  \centering
  \begin{tabular}{lll}
    \toprule
    \cmidrule(r){1-2}
    Seq.Len., (sec.)     & Test Accuracy on LSTM-2ReLU ($\%$) \\
    \midrule
    20   & 95.6     \\
    10   & 92.3      \\
    5    & 86.7  \\
    3    & 73.5  \\
    \bottomrule
  \end{tabular}
  \label{tab:seqlen}
\end{table}

\subsection{Result and Discussion}
\label{sec:heading}

Eight different RNNs were evaluated and their performances of each model are summarized in Table~\ref{tab:result}. These models vary in terms of 3 aspects: the number of hidden layers: both LSTM and ReLU layers; the number of LSTM cells in each hidden layer. Also following \cite{li2015constructing}, the performance of stacking layers is also analyzed with model  (LSTM-128H-2ReLU)$\times$2 and  (LSTM-128H-2ReLU)$\times$3. The relative prediction time needed for the same number of test cases is also given in the table. 

It can be seen that the test accuracy increases as the network becomes deeper. The accuracy can be improved by adding either LSTM layer or ReLU layer, and adding LSTM layer benefits more than adding ReLU layer. Increasing the number of cells in each layer can also improve the accuracy. However, increasing the number of layers or number of cells can also increase the prediction time, which is not good for our general purpose. Comparing the network LSTM-128H-2ReLU and LSTM-128H-1ReLU, which we consider the effect of reducing the size of network, the test accuracy greatly drops from 86.7$\%$ to 78.5$\%$. This is probably because the number of parameters in the network is lower than the minimum requirement of modeling the experiment cases. Besides, the comparison between adding single hidden layers and stacking networks (e.g. 2LSTM-128H-2ReLU and (LSTM-128H-2ReLU)$\times$2) shows that 2 stacked network doesn't outperform the 2 intermediate LSTM layer network. However, the author considers that it cannot lead to a solid conclusion that stacking deep networks is not as beneficial as intermediate hidden layers. The result of this paper may be affected by the total number of training data. Further studies can be performed if more data is available. 

\begin{table}[h]
 \caption{Two-phase flow regime classification results}
  \centering
  \begin{tabular}{lll}
    \toprule
    \cmidrule(r){1-2}
    Network Descriptions   & Test Accuracy ($\%$) & Relative prediction time\\
    \midrule
    LSTM-128H-2ReLU   & 86.7 & 1.00\\
    LSTM-256H-2ReLU   & 88.3 & 1.56    \\
    2LSTM-128H-2ReLU    & 91.7 & 2.18 \\
    3LSTM-128H-2ReLU    & 92.3  & 2.81 \\
    LSTM-128H-1ReLU    & 78.5  & 0.84\\
    LSTM-128H-3ReLU    & 87.1  & 1.32\\
    (LSTM-128H-2ReLU)$\times$2 & 90.2 & 4.28\\
    (LSTM-128H-2ReLU)$\times$3 & 91.1 & 6.11\\    
    \bottomrule
  \end{tabular}
  \label{tab:result}
\end{table}

\section{Conclusions and future work}

The paper developed a methodology of using deep-RNNs for flow regime prediction that can achieve both accuracy and fast response. The method could be extended to the prediction with any time-series database that records the levels and the variations of two-phase parameters, such as void fraction and interfacial area concentration, over time.

\bibliographystyle{unsrt}  
\bibliography{references}

\begin{thebibliography}{10}

\bibitem{mishima1984flow}
K~Mishima and M~Ishii.
\newblock Flow regime transition criteria for upward two-phase flow in vertical
  tubes.
\newblock {\em International Journal of Heat and Mass Transfer},
  27(5):723--737, 1984.

\bibitem{mishima1996some}
K~Mishima and T~Hibiki.
\newblock Some characteristics of air-water two-phase flow in small diameter
  vertical tubes.
\newblock {\em International journal of multiphase flow}, 22(4):703--712, 1996.

\bibitem{barnea1983flow}
D~Barnea, Y~Luninski, and Y~Taitel.
\newblock Flow pattern in horizontal and vertical two phase flow in small
  diameter pipes.
\newblock {\em The Canadian Journal of Chemical Engineering}, 61(5):617--620,
  1983.

\bibitem{cortes1995support}
C~Cortes and V~Vapnik.
\newblock Support-vector networks.
\newblock {\em Machine learning}, 20(3):273--297, 1995.

\bibitem{kohonen1990self}
T~Kohonen.
\newblock The self-organizing map.
\newblock {\em Proceedings of the IEEE}, 78(9):1464--1480, 1990.

\bibitem{mi1998vertical}
Y~Mi, M~Ishii, and LH~Tsoukalas.
\newblock Vertical two-phase flow identification using advanced instrumentation
  and neural networks.
\newblock {\em Nuclear Engineering and Design}, 184(2-3):409--420, 1998.

\bibitem{dang2019investigation}
Z~Dang, Y~Zhao, G~Wang, P~Ju, Q~Zhu, X~Yang, R~Bean, and M~Ishii.
\newblock Investigation of the effect of the electrode distance on the
  impedance void meter performance in the two-phase flow measurement.
\newblock {\em Experimental Thermal and Fluid Science}, 101:283--295, 2019.

\bibitem{yunlong2008identification}
Y~Zhou, F~Chen, and B~Sun.
\newblock Identification method of gas-liquid two-phase flow regime based on
  image multi-feature fusion and support vector machine.
\newblock {\em Chinese Journal of Chemical Engineering}, 16(6):832--840, 2008.

\bibitem{hochreiter1997long}
S~Hochreiter and J~Schmidhuber.
\newblock Long short-term memory.
\newblock {\em Neural computation}, 9(8):1735--1780, 1997.

\bibitem{li2015constructing}
X~Li and X~Wu.
\newblock Constructing long short-term memory based deep recurrent neural
  networks for large vocabulary speech recognition.
\newblock In {\em 2015 IEEE International Conference on Acoustics, Speech and
  Signal Processing (ICASSP)}, pages 4520--4524. IEEE, 2015.

\bibitem{graves2013speech}
A~Graves, Ar~Mohamed, and G~Hinton.
\newblock Speech recognition with deep recurrent neural networks.
\newblock In {\em 2013 IEEE international conference on acoustics, speech and
  signal processing}, pages 6645--6649. IEEE, 2013.

\bibitem{pune9203}
C.~Eberle, M.~Ishii, and S~Revankar.
\newblock A review of gamma densitometer design and measurement in two-phase
  flows, pu/ne-92/3.
\newblock Technical report, Purdue University, 1992.

\bibitem{hewitt1978measurement}
G~Hewitt.
\newblock Measurement of two phase flow parameters.
\newblock {\em Nasa Sti/recon Technical Report A}, 79, 1978.

\bibitem{saxe2013exact}
A~M Saxe, J~L McClelland, and S~Ganguli.
\newblock Exact solutions to the nonlinear dynamics of learning in deep linear
  neural networks.
\newblock {\em arXiv preprint arXiv:1312.6120}, 2013.

\end{thebibliography}

\end{document}